\pdfoutput=1

\documentclass[11pt]{article}

\usepackage[]{emnlp2022}

\usepackage{times}
\usepackage{latexsym}
\usepackage{pgfplots}
\usepackage{lipsum} 
\usepackage{multirow}
\usepackage{colortbl}
\usepackage{booktabs}
\usepackage{amssymb}
\usepackage{subcaption}
\usepackage{tikz}
\usetikzlibrary{shapes,arrows,positioning}

\usepackage{array,booktabs,arydshln,xcolor}
\usepackage{verbatim}
\usepackage{url}
\usepackage{graphicx}
\usepackage{algorithm}
\usepackage{algpseudocode}
\usepackage{amsmath}
\algnewcommand{\Inputs}[1]{%
  \State \textbf{Inputs:}
  \Statex \hspace*{\algorithmicindent}\parbox[t]{.8\linewidth}{\raggedright #1}
}
\algnewcommand{\Initialize}[1]{
  \State \textbf{Initialize:}
  \Statex \hspace*{\algorithmicindent}\parbox[t]{.8\linewidth}{\raggedright #1}
}

\usepackage[T1]{fontenc}

\usepackage[utf8]{inputenc}

\usepackage{microtype}
\usepackage{diagbox}

\title{Learning to Infer from Unlabeled Data: A Semi-supervised Learning Approach for Robust Natural Language Inference}

\author{Mobashir Sadat \mbox{   }\mbox{   }\mbox{   }\mbox{   }\mbox{   }\mbox{   }   Cornelia Caragea\\
  Computer Science \\
  University of Illinois Chicago \\
  \texttt{msadat3@uic.edu \mbox{   }\mbox{   }  cornelia@uic.edu} 
  }

\begin{document}
\maketitle
\begin{abstract}
Natural Language Inference (NLI) or Recognizing Textual Entailment (RTE) aims at predicting the relation between a pair of sentences (premise and hypothesis) as entailment, contradiction or semantic independence. Although deep learning models have shown promising performance for NLI in recent years, they rely on large scale expensive human-annotated datasets. Semi-supervised learning (SSL) is a popular technique for reducing the reliance on human annotation by leveraging unlabeled data for training. However, despite its substantial success on single sentence classification tasks where the challenge in making use of unlabeled data is to assign ``good enough'' pseudo-labels, for NLI tasks, the nature of unlabeled data is more complex: one of the sentences in the pair (usually the hypothesis) along with the class label are missing from the data and require human annotations, which makes SSL for NLI more challenging. In this paper, we propose a novel way to incorporate unlabeled data in SSL for NLI where we use a conditional language model, BART to generate the hypotheses for the unlabeled sentences (used as premises). Our experiments show that our SSL framework successfully exploits unlabeled data and substantially improves the performance of four NLI datasets in low-resource settings. We release our code at: \url{https://github.com/msadat3/SSL_for_NLI}.
\end{abstract}

\section{Introduction}
Natural Language Inference (NLI) or Recognizing Textual Entailment (RTE) is the task of predicting whether a hypothesis entails, contradicts or is neutral to a given premise.  It is widely used as a benchmark for evaluating Natural Language Understanding (NLU) which plays a key role in many Natural Language Processing tasks such as text summarization, machine translation and sentiment analysis. In addition to serving as a benchmark for NLU, NLI has aided in improving the performance in downstream tasks such as fake news detection \cite{sadeghi2022fake} and fact verification \cite{martin2022facter}.

\tikzstyle{premise} = [rectangle,draw,fill=blue!20,text width=10em, text centered, rounded corners, minimum height = 4em]
\tikzstyle{entailment_gen} = [rectangle,draw,fill=green!20,text width=5.5em, text centered, minimum height = 4em]
\tikzstyle{contradiction_gen} = [rectangle,draw,fill=red!20,text width=5.5em, text centered, minimum height = 4em]
\tikzstyle{neutral_gen} = [rectangle,draw,fill=gray!20,text width=5.5em, text centered, minimum height = 4em]

\tikzstyle{entailment_hypo} = [rectangle,draw,fill=green!30,text width=10em, text centered, rounded corners, minimum height = 4em]
\tikzstyle{contradiction_hypo} = [rectangle,draw,fill=red!30,text width=10em, text centered, rounded corners, minimum height = 4em]
\tikzstyle{neutral_hypo} = [rectangle,draw,fill=gray!30,text width=10em, text centered, rounded corners, minimum height = 4em]

\tikzstyle{labeling} =[ellipse,draw,text width=6em, text centered, minimum height = 4em]

\tikzstyle{edge_style} = [draw=black, line width=1]
\begin{figure*}[t]
    \centering
    \begin{tikzpicture}[node distance=2mm and 10mm]
    \small
    \node [premise](p1){I'm sure I shall be only too delighted to make myself useful, I responded.};
    \node[labeling, above left=7mm and -15mm of p1](l_ul){Unlabeled premise};
    \node[labeling, below left=7mm and -15mm of p1](l_clm){Conditional language models};
    \node[neutral_gen, right=of p1](n_gen){Neutral Generation};
    \node[entailment_gen, above=of n_gen](e_gen){Entailment Generation}[above right=0.7cm and 4cm of p1];
    \node[contradiction_gen, below=of n_gen](c_gen){Contradiction Generation};
    \node[neutral_hypo, right=of n_gen](n_hypo){I volunteered to help out for two days.};
    \node[entailment_hypo, right=of e_gen](e_hypo){I am happy to provide assistance.};
    \node[contradiction_hypo, right=of c_gen](c_hypo){I am unhappy to make myself useful.};
    \node[labeling, right=of n_hypo](l_sh){Generated hypotheses};
    
    \draw [->] (p1) edge (n_gen);
    \draw [->] (p1) edge (c_gen);
    \draw [->] (p1) edge (e_gen);
    \draw [->] (e_gen) edge (e_hypo);
    \draw [->] (c_gen) edge (c_hypo);
    \draw [->] (n_gen) edge (n_hypo);
    \draw [dotted] (l_ul) edge (p1);
    \draw [dotted] (l_sh) edge (e_hypo);
    \draw [dotted] (l_sh) edge (c_hypo);
    \draw [dotted] (l_sh) edge (n_hypo);
    \draw [dotted] (l_clm) edge (n_gen);
    \draw [dotted] (l_clm) edge (c_gen);
    \draw [dotted] (l_clm) edge (e_gen);

    \end{tikzpicture}
    \caption{An example of synthetic hypotheses generation from an unlabeled premise.}
    \label{fig:hypo_gen_example}
\end{figure*}
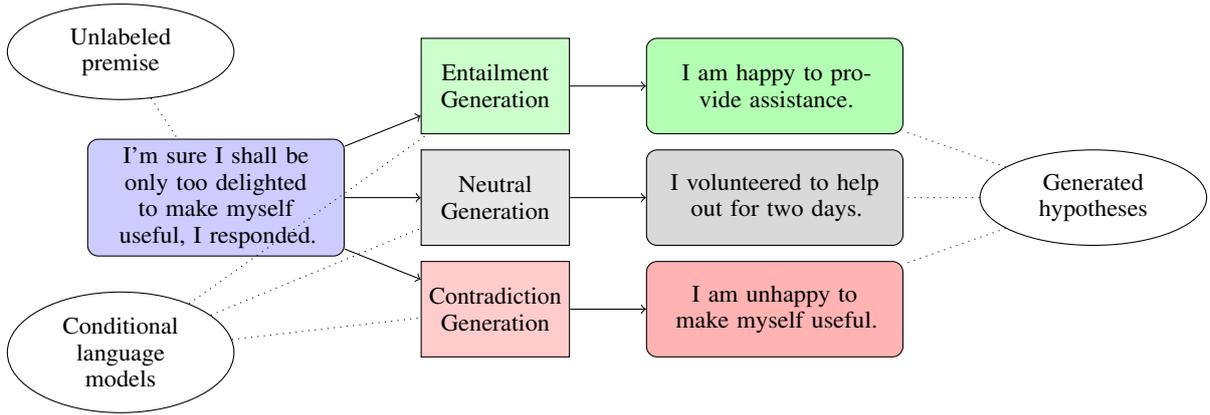

In recent years, deep learning based approaches \cite{chen2016enhanced, devlin-etal-2019-bert, liu2019roberta} have shown promising performance for NLI due to their superior ability to extract deep semantic features. However, despite their success, one of the key challenges of the deep learning based models is that they require large scale human annotated datasets to perform well. Consequently, earlier NLI datasets such as SICK \cite{marelli-etal-2014-sick} and RTE \cite{dagan2005pascal}, while being instrumental in the progress of NLI research, are not suitable for training these models due to their small size. To address this, large scale datasets such as SNLI \cite{bowman-etal-2015-large}, MNLI \cite{N18-1101}, and ANLI \cite{nie-etal-2020-adversarial} have been proposed. The general annotation protocol for these datasets involves curating the premises from a pre-existing source and then employing human crowdworkers to write the hypotheses and assign their class labels. Given the large number of annotated samples needed for a dataset to be suitable for the deep learning models, their construction requires a significant amount of human effort. Consequently, creating new NLI datasets capturing unique linguistic properties of different domains/time/demography becomes cumbersome and in some cases impossible. Therefore, it is necessary to reduce the reliance on manually annotated data in training deep learning models for NLI.

To this end, we seek to harness unlabeled data by adopting semi-supervised learning (SSL) for low-resource NLI datasets which could potentially improve the performance without the necessity of any additional human effort. SSL is a widely used method for automatically assigning pseudo-labels to unlabeled data to incorporate them into model training \cite{xie2020self, becker-etal-2013-avaya, NEURIPS2021_c1fea270}. However, unlike single sentence classification tasks, where the challenge in making use of unlabeled data is to assign ``good enough'' pseudo-labels, as described above, the nature of unlabeled data is more complex for NLI: one of the sentences in the pair (usually the hypothesis) along with the class label are missing from the data and require human annotations. Therefore, in order to leverage unlabeled data for NLI, the unavailability of both hypotheses and class labels need to be tackled. Thus, employing SSL is much more challenging for NLI than other tasks. Potential approaches for addressing this challenge include selecting other unlabeled sentences as hypotheses randomly or identifying them based on similarity with the unlabeled premises. However, these approaches can either result in all neutral samples and fail to provide the necessary coverage to the other classes or can be computationally too expensive. As a result, the unavailability of hypotheses for unlabeled premises remains as the key bottleneck in exploring SSL for NLI.

In this paper, we propose to address this bottleneck with a novel approach for incorporating unlabeled data into training low-resource NLI models where we generate the hypotheses for the unlabeled premises with a state-of-the-art conditional language model, BART \cite{lewis-etal-2020-bart}. Specifically, for each unlabeled premise, one hypothesis is generated corresponding to each class in the labeled dataset. As a result, our proposed method guarantees the coverage of all classes during training. We can see an example of our hypothesis generation method in Figure \ref{fig:hypo_gen_example}. Then, we develop an SSL framework based on our proposed strategy for leveraging unlabeled data for NLI where we adopt iterative self-training to gradually accumulate useful pseudo-labeled examples for training. Our experiments show that the proposed SSL framework can improve both in-domain and out-of-domain performance on four NLI datasets in low-resource settings. Our contributions are as follows:

\begin{itemize}
    \item We propose a novel approach for incorporating unlabeled data in training models on low-resource NLI datasets. To our knowledge, we propose the first ever SSL framework for NLI based on this approach.
   \item We thoroughly evaluate our proposed framework using four NLI datasets in low-resource settings and show that both in-domain and out-of-domain performances improve substantially illustrating that we successfully leverage unlabeled data for NLI. 
   
   \item We perform a comprehensive analysis of our framework to understand whether it can improve the robustness of NLI models. Our findings suggest that the overall robustness of the models improves by a considerable margin when they are trained using our framework.
\end{itemize}

\section{A New SSL Framework for NLI}

This section details our proposed SSL for NLI framework which consists of two components: hypothesis generation and self-training. An overview of our framework can be seen in Algorithm \ref{alg:self_training}.

\paragraph{Problem Formulation} Let $D^{l} = \{(p^{l}_{i}, h^{l}_{i}, y^{l}_{i})\}_{i=1,...,n}$ be our labeled set of size $n$ where $p^{l}_{i}$, $h^{l}_{i}$, and $y^{l}_{i}$ represent the premise, hypothesis and label of sample $i$ and let $C$ be the set of classes in $D^l$. Given $D^l$ and a large set of unlabeled premises $\{p^{u}_{i}\}_{i=1,...,m}$ of size $m$, our SSL framework has two objectives. First, it creates a synthetically annotated dataset $D^{syn}$ using our proposed hypothesis generation method. Next, it employs self-training to iteratively assign pseudo-labels and select examples from $D^{syn}$ which are of high quality and puts them in our pseudo-labeled set $D^p$ to be used for training in addition to the labeled dataset $D^l$.

\algnewcommand\algorithmicforeach{\textbf{for each}}
\algdef{S}[FOR]{ForEach}[1]{\algorithmicforeach\ #1\ \algorithmicdo}

\begin{algorithm*}
\small
\caption{Semi-supervised NLI}\label{alg:self_training}
\begin{algorithmic}[1]
\Require 
\Statex Labeled samples $D^{l} = \{(p^{l}_{i}, h^{l}_{i}, y^{l}_{i})\}_{i=1,..n}$;

\Statex Unlabeled premises $P^{u} = \{(p^{u}_{i})\}_{i=1,..m}$;

\State Initialize synthetically labeled set, $D^{syn} \gets \emptyset$

\ForEach {$c \in C$}: \Comment{C is the set of classes in $D^l$}
\State Fine-tune BART to generate hypotheses conditioned on the premises for class $c$ on corresponding examples from $D^l$
\State Generate hypotheses for class $c$ for each unlabeled premise $p^{u}_{i}$ using the fine-tuned BART model
\State Add each unlabeled premise, generated hypothesis and class $c$ as a synthetically labeled example to $D^{syn}$
\EndFor

\State Initialize pseudo-labeled set, $D^{p} \gets \emptyset$ 
\State Initialize current iteration $k \gets 0$
\While{$k \neq K$ \textbf{and} $D^{syn} \neq \emptyset$} \Comment{K is a pre-defined maximum iteration.}

\State Train classification model, $\theta_{k}^{clf}$ on a combination of $D^l$ and $D^p$

\State $D^{syn}_k \gets S$ examples randomly sampled from $D^{s}$

\State Predict the labels of the samples in $D^{syn}_k$ and get the model confidences

\State Apply filters on the samples in $D^{syn}_k$ using Equation \ref{eqn:selection} and put the selected samples in $D^{p}_{k}$

\State Add new pseudo-labeled examples to the pseudo-labeled set, $D^{p} \gets D^{p} \cup D^{p}_k$ 

\State Remove new pseudo-labeled examples from the synthetically labeled set, $D^{syn} \gets D^{syn} \setminus D^{p}_k$  

\State Increment iteration, $k \gets k + 1$ 
\EndWhile

\end{algorithmic}
\end{algorithm*}

\subsection{Hypothesis Generation}
\label{sec:hypo_gen}
In order to harness unlabeled data for NLI, for each unlabeled premise, we propose to generate one hypothesis corresponding to each class using a conditional language model to ensure both class coverage and computational efficiency (compared to exhaustively searching for related sentence pairs in a large corpus). For our experiments, we use BART \cite{lewis-etal-2020-bart} as the conditional language model which was pre-trained by learning to reconstruct text corrupted in different manners such as token masking, token deletion, sentence permutation etc. However, any conditional language model that can generate text can be used in its place in our proposed framework.

For each class, we first select its corresponding premise-hypothesis pairs from $D^l$ and then fine-tune the pre-trained BART model using cross-entropy loss where the premises from the selected pairs are used as the source texts and their hypotheses as the targets. We denote the fine-tuned model for each class $c$ as $BART^c$. The synthetically labeled dataset $D^{syn}$ is then created as follows:
\begin{multline}
    D^{syn} = [\{(p^{syn}_{i} = p^{u}_{i}, h^{syn}_{i} = BART^{c}(p^{u}_{i}), \\ y^{syn}_{i} = c)\}_{i=1,..m}]_{c \in C}
\end{multline}
That is, if there are three classes --- entailment, contradiction and neutral in the dataset, each unlabeled premise $p^{u}_{i}$ is paired with three synthetic hypotheses generated with the BART models corresponding to these classes. We also assign the respective class of the model used to generate the hypothesis as the initial synthetic label for each premise-hypothesis pair.  

\subsection{Self-training}
\label{Sec:ST}
Self-training is an iterative algorithm which assigns pseudo-labels to unlabeled data at each iteration and selects a subset of them based on some quality assurance measures to be used as additional training data in the subsequent iterations. 

Due to the sole reliance on the generative models for assigning the synthetic labels for the premise-hypothesis pairs, our synthetically annotated dataset $D^{syn}$ can contain harmful noise and significantly affect classification performance if they are directly used as training data without any quality assurance measures. Therefore, we make use of self-training to iteratively curate high quality samples from $D^{syn}$ and put them into our pseudo-labeled set $D^{p}$. To this end, at each iteration $k$, first, we train a classifier $\theta_{k}^{clf}$ on a combination of labeled training set $D^l$ and pseudo-labeled training set $D^{p}$ if it is non-empty (i.e., if $k > 1$). Next, to provide equal coverage to all classes, we randomly sample a balanced subset of $S$ examples from our synthetically labeled set $D^{syn}$ for pseudo-labeling. We denote this sampled synthetic subset as $D^{syn}_k$. The pseudo-label $y_{i}^{p}$ for each sample $i$ in this subset is then assigned as follows:
\begin{equation}
    y_{i}^{p} = \arg \max_{c \in C} {\theta_{k}^{clf}}(c | p_{i}^{syn}, h_{i}^{syn})
\end{equation}
Here, ${\theta_{k}^{clf}}$ predicts the probability distribution of each sample over $C$ classes conditioned on the premise $p_{i}^{syn}$ and hypothesis $h_{i}^{syn}$.

\paragraph{Quality Assurance}
To ensure data quality, we apply two filters to our pseudo-labeled data. First, following the traditional approach for quality assurance of pseudo-labeled data in SSL, the examples for which the pseudo-label is predicted with a confidence score lower than a pre-defined threshold $\tau$ are filtered out. Next, since we have two sets of labels for the unlabeled samples, i.e., synthetically assigned label $y_{i}^{syn}$ (assigned through the BART model) and predicted pseudo-label $y_{i}^{p}$ (assigned by the classifier) for each sample $i$, we enforce their consistency as an additional data quality measure. Therefore, we filter out the examples for which the predicted pseudo-label $y_{i}^{p}$ and the initially assigned synthetic label $y_{i}^{syn}$ do not match. More formally, based on our two filters, we select the pseudo-labeled examples for iteration $k$ as follows:
\begin{multline}
\label{eqn:selection}
 D_{k}^{p} = \{(p_{i}^{syn}, h_{i}^{syn}, y_{i}^{syn}) : y_{i}^{syn} = y_{i}^{p}, \\ {\theta_{k}^{clf}}( y_{i}^{p}| p_{i}^{syn}, h_{i}^{syn}) \ge \tau \}_{i=1,..S} 
\end{multline}
After selecting the pseudo-labeled examples from the current iteration, we add them to $D^{p}$, our pseudo-labeled set which is accumulated across the iterations:
\begin{equation}
    D^{p} = D^{p} \cup  D_{k}^{p}
\end{equation}
The self-training iterations are run until $D^{syn}$ is empty, i.e., we run out of synthetic examples or a pre-defined number of iterations, $K$ is reached. The model from the iteration showing the best development score is used to evaluate the test sets. 

Since we follow the traditional practice of self-training by training the classification model on a combination of labeled and pseudo-labeled data in each iteration, we denote this self-training setup as \textbf{VST} which stands for Vanilla Self-Training.

\paragraph{De-biased Self-training} Given that the hypotheses in our pseudo-labeled datasets are machine generated, we hypothesize that they can introduce some unnecessary bias to the classification models despite our strict quality assurance measures for curating them. Therefore, we explore a self-training setup where we first train the classification model on pseudo-labeled data and then train it on human annotated labeled data in each iteration to reduce the introduced bias. This model is denoted by \textsc{\textbf{DBST}} or de-biased self-training.  

\paragraph{Noised Self-training} Inspired by recent work on self-training for image classification \cite{xie2020self} which shows that perturbing an input image by introducing controlled noise/data augmentation after assigning the pseudo-labels can significantly improve the classification performance, we explore perturbed / noised versions of our self-training methods. Specifically, after selecting the pseudo-labeled data using Equation \ref{eqn:selection}, we employ back-translation \cite{yu2018fast} to replace the original premise and hypothesis in each sample with their augmented versions. The perturbed versions of our vanilla and de-biased self-training setups are denoted as \textsc{\textbf{VST + N}} and \textsc{\textbf{DBST + N}}.
\section{Datasets}
\label{sec:datasets}
We use the following datasets for our experiments. 

\vspace{1mm}
\noindent \textbf{\textsc{SICK}} \cite{marelli-etal-2014-sick} \quad SICK is a dataset containing  4,500 sentence pairs in its training set. We use the 8K ImageFlickr data set\footnote{\url{https://www.kaggle.com/adityajn105/flickr8k/activity}} to extract unlabeled premises. This dataset was one of the sources for deriving \textsc{SICK}.

\noindent \textbf{\textsc{RTE}} \cite{wang-etal-2018-glue} \quad This dataset contains $\approx 2.5K$ examples and was created by combining the RTE1 \cite{dagan2005pascal}, RTE2 \cite{haim2006second}, RTE3 \cite{giampiccolo2007third} and RTE5 \cite{bentivogli2009fifth} datasets. The premises for RTE were extracted from Wikipedia and news sources. Therefore, we also extract sentences from Wikipedia and the CNNDM \cite{nallapati-etal-2016-abstractive} dataset to be used as unlabeled premises. Since, the test set of \textsc{RTE} is not publicly available, we use its development set as the test set and randomly sample a small subset from the training set to be used as the development set.

\noindent \textbf{\textsc{MNLI - 6K}} \cite{N18-1101} \quad To simulate a low-resource environment, we randomly sample 6,000 examples from the training set of MNLI. From the rest of the training set, we select the premises which do not occur in the sampled 6,000 examples to be used as unlabeled data. Similar to \textsc{RTE}, we use the development set of MNLI as the test set and sample a small subset of examples from the training set to be used as development set.

\noindent \textbf{\textsc{SNLI - 6K}} \cite{bowman-etal-2015-large} \quad We created SNLI - 6K in a similar fashion as \textsc{MNLI - 6K}.

\section{Baselines}
Given that we are the first to explore NLI in a low-resource setting, there are no existing methods that we can use as baselines, which are specifically tailored for NLI when human annotated data is limited. Thus, most of our baselines are adopted from methods which have been successful for other NLP tasks in low-resource settings. Specifically, we compare the performance of our SSL framework with three types of baselines.

\paragraph{\textsc{BERT} \cite{devlin-etal-2019-bert}:} This is a baseline in which a BERT model is fine-tuned only on the available human-annotated data.
\paragraph{\textsc{Data Augmentation (DA)}} We compare the performance of our SSL framework with three data augmentation baselines. We augment both premise and hypothesis in each example in the labeled training set and combine the original labeled set with its augmented version. In other words, the size of the labeled dataset is doubled by adding an augmented version of each original example. We use the following data augmentation methods. 

\noindent \textbf{(a)} \textsc{Back-translation (BT)} \cite{yu2018fast} \quad The pair (premise and hypothesis) in each example in the labeled training set is translated to French and then translated back to English using machine translation models to get their paraphrased versions.
    
\noindent \textbf{(b)} \textsc{Synonym Replacement (SR)} \cite{kolomiyets-etal-2011-model} \quad Randomly chosen tokens from both premise and hypothesis are replaced with their synonyms using WordNet \cite{wordnet}.
    
\noindent \textbf{(c)} \textsc{Conditional Masked Language Modeling (C-MLM)} \quad Inspired by CBERT \cite{wu2019conditional} where some randomly chosen positions in the input text are masked and a conditional masked language model (C-MLM) is used to predict the tokens in those positions based on both the context and the class label to get the augmented versions of the original input text, we formulate a similar DA method for NLI. Specifically, the premise, hypothesis and label of each sample is combined using class-specific templates that we describe in Appendix \ref{appndx: C_mlm}. We then randomly mask a subset of common tokens in both premise and hypothesis and use the C-MLM model to predict the tokens at those positions to get their augmented versions.s

\setlength\dashlinedash{0.2pt}
\setlength\dashlinegap{1.5pt}
\setlength\arrayrulewidth{0.3pt}
\begin{table*}[t]
\centering
\small
  \begin{tabular}{ l c c c c c}
    \toprule
{\bf Approach} &  {\bf \textsc{RTE}} & {\bf \textsc{SICK}} & {\bf \textsc{SNLI-6K}} & {\bf \textsc{MNLI-6K$_{m}$}} & {\bf \textsc{MNLI-6K$_{mm}$}}\\
  \midrule
    \textsc{BERT} & $60.90 \pm 2.99$ & $84.63 \pm 0.66$ & $78.47 \pm 0.26$ & $68.76 \pm 0.56$ & $70.05 \pm 0.73$\\
    \hdashline
    \textsc{BT}  & $62.27 \pm 1.69$ & $84.48 \pm 0.47$ & $78.39 \pm 0.35$ & $69.52 \pm 0.51$  & $71.20 \pm 0.26$ \\
    \textsc{SR}  & $60.05 \pm 1.17$ & $84.11 \pm 0.75$ & $78.33 \pm 0.22$ & $68.29 \pm 0.11$  & $68.35 \pm 0.23$ \\
    \textsc{C-MLM} & $62.42 \pm 0.19$ & $84.74 \pm 0.36$ & $78.85 \pm 0.29$ & $69.38 \pm 0.56$  & $70.79 \pm 0.32$ \\

   \hdashline
    \textsc{ST - RH}  & $63.28 \pm 0.85$ & $84.93 \pm 0.18$  & $78.31 \pm 0.52$ & $68.58 \pm 0.36$  & $70.21 \pm 0.64$ \\
    \textsc{UDA}   & $58.31 \pm 2.17$ & $84.40 \pm 0.57$ & $78.78 \pm 0.61$ & $69.46 \pm 0.08$ & $71.05 \pm 0.42$  \\
    \hdashline
    \textsc{VST} & $64.18 \pm 2.10$ & $85.02 \pm 0.32$ & $79.46^{*} \pm 0.25$  & $71.57^{*} \pm 0.34$ &  $73.03^{*} \pm 0.47$  \\
    \mbox{{\color{white}{xxx}}+}\textsc{N} & $66.31 \pm 2.66$ & $84.64 \pm 0.44$ & $79.01 \pm 0.07$ & $71.13^{*} \pm 0.62$ & $72.46^{*} \pm 0.14$ \\
    \hdashline
    \textsc{DBST}   & $67.98^{*} \pm 1.78$ & $\textbf{85.77} \pm \textbf{0.06}$ & $80.04^{*} \pm 0.35$  & $72.22^{*} \pm 0.61$ & $74.02^{*} \pm 0.29$ \\
    \mbox{{\color{white}{xxx}}+}\textsc{N}   & $\textbf{68.32}^{*} \pm \textbf{2.03}$ & $85.51 \pm 0.33$ & $\textbf{80.31}^{*} \pm \textbf{0.21}$ & $\textbf{72.99}^{*} \pm \textbf{0.10}$ & $\textbf{74.48}^{*} \pm$ $\textbf{0.22}$ \\
    \bottomrule
  \end{tabular}
  \caption{The average and standard deviation of the Macro F1 (\%) from three different runs of different approaches on our selected datasets. Here, RH and N stands for Random Hypothesis and Noise, respectively. Best scores are in bold. An asterisk indicates a statistically significant difference with BERT according to a paired T-test with $\alpha = 0.05$.}
    \label{table:results comparison}
\end{table*}

\paragraph{\textsc{Unlabeled Data Exploitation}:}
We use two baselines that leverage unlabeled data using methods different from our SSL framework to evaluate the efficacy of our proposed method for leveraging unlabeled data. 

\noindent \textbf{(a)} \textsc{Unsupervised Data Augmentation (UDA)} \cite{xie2020unsupervised} \quad   This is a semi-supervised learning framework where traditional cross-entropy loss on the labeled data is combined with a consistency loss which is aimed at minimizing the distance between the probability distributions predicted by the model for an unlabeled sample and an augmented version of the unlabeled sample. The quality assurance of unlabeled data is done during training based on a pre-defined confidence threshold. We use back-translation to augment both premise and hypotheses of each unlabeled sample.

\noindent \textbf{(b)} \textsc{Self-training with Random Hypothesis (ST - RH)} \quad This is a self-training approach which is similar to \textsc{DBST} except it uses randomly chosen sentences as the hypotheses for unlabeled premises instead of synthetically generating them.

\section{Main Experiments \& Results}
Our main experiments and results are described in this section. We run each of our self-training and baseline experiments three times and report the average and standard deviation of their Macro F1 scores in Table \ref{table:results comparison}. Our implementation details can be found in Appendix \ref{apndx:implementation_details}.

\paragraph{\textsc{DBST} vs \textsc{BERT}}  To understand the effectiveness of our SSL framework, we compare the performance of our \textsc{DBST} model with the baseline BERT model. The results show that \textsc{DBST} substantially improves the performance over \textsc{BERT} for all four datasets. For example, we can see an increase of $7.08\%$ and $3.97\%$ in Macro F1 by \textsc{DBST} over \textsc{BERT} on \textsc{RTE} and \textsc{MNLI-6K$_{mm}$}, respectively. Our qualitative analysis of the synthetically-labeled data suggests that the BART models used in our framework are able to generate meaningful hypotheses for each premise (see Table \ref{table:synthetic_examples} in Appendix \ref{appndx:synthetic_examples} for examples). Furthermore, since we generate one hypothesis for each unlabeled premise corresponding to each class, our proposed method guarantees the coverage of all classes. Consequently, we are able to expose the models to high quality premise-hypothesis pairs derived from unlabeled data which enables \textsc{DBST} to show consistent improvements proving that our SSL framework successfully harnesses unlabeled data for NLI. 

\paragraph{\textsc{DBST} vs DA Baselines} Next, we evaluate whether our SSL framework can improve the performance over prior methods which handle low-resource scenarios without leveraging unlabeled data. To this end, we compare the performance of the \textsc{DBST} model with our DA baselines. We can see in Table \ref{table:results comparison} that \textsc{DBST} outperforms all our DA baselines by a significant margin. Moreover, in general, the DA methods only minimally improve the performance over BERT. For example, the best performing DA method for \textsc{SICK}, \textsc{C-MLM} shows an improvement of $0.11\%$ in Macro F1 score over BERT. Similarly, the improvement shown by \textsc{BT} for \textsc{MNLI-6K$_{m}$} is only $0.76\%$. These results indicate that DA methods fail to introduce enough diversity in the augmented sentences which can help improve the models' ability in recognizing semantic relations between premise-hypothesis pairs. In contrast, harnessing unlabeled data with our SSL framework provides us with linguistically diverse patterns beneficial in improving the performance.

\paragraph{\textsc{DBST} vs Alternative Methods for Leveraging Unlabeled Data} We now
compare the performance of the \textsc{DBST} model with \textsc{ST - RH} and \textsc{UDA} to evaluate our proposed strategy for leveraging unlabeled data for NLI. Both of these models aims at exploiting unlabeled data with a strategy different than ours. \textsc{ST - RH} randomly chooses other unlabeled sentences for each premise instead of synthetically generating them while \textsc{UDA} employs an additional consistency loss term between the original and perturbed version of unlabeled data. The results in Table \ref{table:results comparison} show that \textsc{DBST} outperforms both of these models. Although in \textsc{UDA}, we use the same synthetic hypothesis generation strategy as in \textsc{DBST}, the results indicate that simply improving the consistency between the predictions made for unlabeled data and their perturbed version is not enough to improve the performance. Comparing \textsc{DBST} with \textsc{ST - RH}, we find that almost all unlabeled examples are assigned a \textsc{Neutral} pseudo-label in \textsc{ST - RH} as we hypothesized. Therefore, unlike our proposed strategy for leveraging unlabeled data, randomly choosing unlabeled sentences as the hypotheses fails to provide enough coverage to all relevant classes which results in self-training not being able to perform well. 

\setlength\dashlinedash{0.2pt}
\setlength\dashlinegap{1.5pt}
\setlength\arrayrulewidth{0.3pt}
\begin{table*}[t]
\centering
\small
  \begin{tabular}{l l c c c c c}
    \toprule
 {\bf Model} & {\bf \backslashbox{Train}{Test}} &  {\bf \textsc{RTE}} & {\bf \textsc{SICK}} & {\bf \textsc{SNLI-6K}} & {\bf \textsc{MNLI-6K$_{m}$}} & {\bf \textsc{MNLI-6K$_{mm}$}}\\
  \midrule
     \textsc{BERT} & {\bf \textsc{RTE}} & - & $46.08 \pm 11.51$ & $60.80 \pm 3.95$ & $61.58 \pm 2.70$ & $35.75 \pm 0.49$ \\
     \textsc{DBST + N} & & - & $\textbf{54.53} \pm \textbf{5.29}$ & $\textbf{69.05} \pm \textbf{1.33}$ & $\textbf{68.39} \pm \textbf{0.69}$ & $\textbf{39.48} \pm \textbf{0.50}$ \\
    \hdashline
    \textsc{BERT} & {\bf \textsc{SICK}} & $49.30 \pm 6.25$ & - & $37.10 \pm 2.74$ & $43.25 \pm 6.96$ & $45.80 \pm 9.36$  \\
    \textsc{DBST + N} & & $48.22 \pm 3.32$ & - & $\textbf{42.19} \pm \textbf{1.13}$ & $\textbf{45.83} \pm \textbf{2.84}$ & $\textbf{50.30} \pm \textbf{3.66}$ \\
    \hdashline
    \textsc{BERT} &  {\bf \textsc{SNLI-6K}} & $56.72 \pm 0.04$ & $44.34 \pm 1.55$ & - & $54.92 \pm 1.20$ & $57.10 \pm 1.44$\\
    \textsc{DBST + N} & & $\textbf{58.21} \pm \textbf{0.55}$ & $\textbf{47.15} \pm \textbf{1.43}$ & - & $\textbf{59.00} \pm \textbf{0.19}$ & $\textbf{61.88} \pm \textbf{0.43}$  \\
    \hdashline
     \textsc{BERT} & {\bf \textsc{MNLI-6K}} & $64.51 \pm 1.70$ & $60.85 \pm 1.75$ & $61.35 \pm 2.40$ & - & - \\
    \textsc{DBST + N} & & $\textbf{64.93} \pm \textbf{0.84}$ & $54.36 \pm 5.62$ & $\textbf{66.44} \pm \textbf{1.25}$ & - & - \\
    \bottomrule
  \end{tabular}
  \caption{Out-of-domain (OOD) Macro F1 (\%) scores of the \textsc{DBST + N} models compared with baseline BERT models. The main diagonal is kept blank because it corresponds to in-domain (ID) performances. 
  }
    \label{table:OOD_results}
\end{table*}

\setlength\dashlinedash{0.2pt}
\setlength\dashlinegap{1.5pt}
\setlength\arrayrulewidth{0.3pt}
\begin{table*}[t]
\centering
\resizebox{\textwidth}!{
  \begin{tabular}{l l c c c c c c c c c c c}
    \toprule
& &  \multicolumn{3}{c}{\bf Competence Test} & \multicolumn{6}{c}{\bf Distraction Test} & \multicolumn{2}{c}{\bf Noise Test}\\
\cmidrule(lr){3-5}
\cmidrule(lr){6-11}
\cmidrule(lr){12-13}
   & & \multicolumn{2}{c}{} & {} & \multicolumn{2}{c}{\bf Word} & \multicolumn{2}{c}{} & \multicolumn{2}{c}{\bf Length} & \multicolumn{2}{c}{\bf Spelling} \\ 
   & & \multicolumn{2}{c}{\bf Antonymy} & {\bf Numerical} & \multicolumn{2}{c}{\bf Overlap} & \multicolumn{2}{c}{\bf Negation} & \multicolumn{2}{c}{\bf Mismatch} & \multicolumn{2}{c}{\bf Error} \\ 
    {\bf Model} & {\bf Dataset} & {\bf Mat} & {\bf Mis} & {\bf Reasoning} & {\bf Mat} & {\bf Mis} & {\bf Mat} & {\bf Mis} & {\bf Mat} & {\bf Mis} & {\bf Mat} & {\bf Mis} \\
   \midrule
      \textsc{BERT} & \textsc{ \bf RTE} & 8.52 & 6.01 & 38.67 & 65.06 & 65.26 & 64.08 & 64.25 & 61.78 & 62.73 & 66.67 & 66.50\\
     \textsc{DBST + N} & & 1.25 & 0.89 & 34.05 & 66.38 & 67.25 & 68.04 & 69.20 & 65.53 & 65.14 & 67.03 & 66.96\\
    \hdashline
     \textsc{BERT} & \textsc{\bf SICK} &  55.26 & 56.50 & 25.80 & 34.48 & 34.36 & 39.67 & 41.57 & 44.42 & 46.62  & 42.09 & 42.49 \\
    \textsc{DBST + N} & & 28.85 & 28.98 & 25.05 & 38.02 & 39.17 & 43. 83 & 46.71 & 47.84 & 51.16 & 43.96 & 44.31\\
    \hdashline
    \textsc{BERT} & \textsc{\bf SNLI-6K} & 18.38 & 16.18 & 28.83 & 42.66 & 44.14 & 51.12 & 52.09 & 54.22 & 55.90 & 49.09 & 49.31 \\
     \textsc{DBST + N} & & 7.06 & 5.89 & 34.56 & 50.92 & 52.87 & 56.55 & 58.56 & 58.00 & 59.70 & 53.28 & 53.43\\
    \hdashline
     \textsc{BERT} & \textsc{\bf MNLI-6K} & 8.03 & 7.69 & 29.23 & 43.66 & 44.87 & 62.25 & 64.02 & 66.29 & 68.75 & 64.80 & 65.32\\
    \textsc{DBST + N} & & 17.72 & 15.43 & 30.32 & 46.32 & 47.36 & 59.06 & 60.43 & 70.43 & 72.71 & 68.35 & 68.65\\
    \bottomrule
  \end{tabular}
  }
  \caption{Stress test accuracies (\%) of the baseline BERT  and \textsc{DBST + N} trained on different datasets.}
    \label{table:stress_test}

\end{table*}

\paragraph{Vanilla vs De-biased} Comparing the performance of \textsc{VST} and \textsc{DBST} in Table \ref{table:results comparison}, we can see that \textsc{DBST} performs better than \textsc{VST} in all four datasets. This corroborates our notion that synthetically generated hypotheses while being useful in improving the performance, can introduce some bias to the model. When we continue training the model on only human-annotated data, this bias gets reduced which leads to better performance.

\paragraph{Noised Model Evaluation} To evaluate whether introducing noise to the pseudo-labeled data by back-translating both premise and hypothesis can further improve the performance, we compare \textsc{VST} and \textsc{DBST} with their noisy counterparts in Table \ref{table:results comparison}. We can see that \textsc{VST + N} shows a lower performance than \textsc{VST}. On the other hand, adding noise to \textsc{DBST} pushes the performance further. For example, on \textsc{MNLI - 6K$_m$}, we see an improvement of $0.77$. \textsc{DBST + N} also improves over \textsc{DBST} for \textsc{RTE} and \textsc{SNLI} but the improvement margin is smaller. This discrepancy in trends shown by the \textsc{VST + N} and \textsc{DBST + N} models indicate that when we noise the input text (i.e., replace it with its augmented versions), in addition to examples which help the model become more robust, some mis-labeled examples (i.e., harmful for the model), are introduced. The additional fine-tuning step in \textsc{DBST + N} on only the original clean data reduces the harmful noise while retaining the robustness introduced by the other useful perturbed examples. As a result, we see an improvement in performance by \textsc{DBST + N} over \textsc{DBST} while the performance of \textsc{VST + N} declines from \textsc{VST} due to a lack of additional fine-tuning step on clean data.
\section{Out-of-domain Results}
To assess whether our SSL framework can improve the out-of-domain (OOD) performance, we evaluate \textsc{DBST + N} and BERT models trained on each dataset using test sets of other datasets. It should be noted that we do not perform any additional training of the models for evaluating the OOD performance. We simply use the models already trained on labeled and/or unlabeled data on the training set of a particular dataset and test them on other datasets. The OOD results are reported in Table \ref{table:OOD_results}.

We can see that in general, \textsc{DBST + N} shows a significantly higher OOD performance than BERT. For example, the performance of \textsc{DBST + N} trained on RTE shows an $8.45\%$ higher Macro F1 score than BERT when they are tested on \textsc{SICK}. Therefore, harnessing unlabeled data with our SSL framework can improve the performance on new domains without any additional annotation effort.

\section{Analysis}
\paragraph{Robustness Analysis} From our experiments and results, it is evident that our SSL framework is effective in improving both in-domain and out-of-domain performances. However, it is not clear if SSL can train more robust NLI models. Therefore, we study the robustness of our models on NLI stress test \cite{naik-etal-2018-stress}. The stress test was created based on the weaknesses of NLI models in various aspects such as word overlap --- models tend to predict a sample to be entailment if there is a large overlap in words even if they are unrelated, negation --- presence of a negation word such as `no' causes the model to predict the sample as contradiction etc. In total, 11 different tests are carried out which are divided into three parts: competence, distraction, and noise. We compare the stress test results for the baseline BERT model and  \textsc{DBST + N} model in Table \ref{table:stress_test}.

\setlength\dashlinedash{0.2pt}
\setlength\dashlinegap{1.5pt}
\setlength\arrayrulewidth{0.3pt}
\begin{table}[t]
\centering
\small
  \begin{tabular}{l c c }
    \toprule
{\bf Approach} & {\bf MNLI-6K$_m$}     & {\bf MNLI-6K$_{mm}$} \\ 
   \midrule
   \textsc{DBST + N} & $73.11$ & $74.66$ \\
    
    \textsc{DBST + N$_{SB}$} & $72.59$ & $74.54$ \\
    \textsc{DBST + N$_{CM}$} & $72.58$ & $74.31$ \\
    \bottomrule
  \end{tabular}
  \caption{The Macro F1 (\%) score of \textsc{DBST + N} compared with \textsc{DBST + N$_{SB}$} and \textsc{DBST + N$_{CM}$}.
  }
  \vspace{-3mm}
    \label{table:ablation}
    
\end{table}

The \textsc{DBST + N} models show better performance than the baseline BERT model on both distraction and noise tests. Therefore, our SSL framework reduces the vulnerability of the models in being distracted by shallow features such as word overlap, negation and mismatch in lengths and improves their ability in making real inferential decisions. Moreover, the models trained with SSL are less prone to making wrong predictions due to the noise caused by spelling mistakes which also helps in improving the robustness.

However, both BERT and \textsc{DBST + N} models fail to show meaningful performance for the competence tests. In general, the accuracy for both models is lower than the random baseline ($50\%$ for RTE since it is a 2-way classification and $33.33\%$ for the other datasets which are 3-way classifications). This indicates that while our proposed SSL framework improves the overall robustness of the models, it cannot address the models' inability to recognize the contradicting relations caused by antonyms and their weakness in reasoning over numeric tokens. Therefore, in our future work we aim at incorporating methods which can make the models more robust in recognizing the antonyms and more capable to numeric reasoning.    

\paragraph{Single BART} In order to understand the necessity of using separate BART models for each class, we perform an experiment on \textsc{MNLI-6K} where we employ a single BART model to generate the hypotheses for all classes in our \textsc{DBST + N} approach. To this end, we append the label at the end of each premise in the labeled training set to be used as the source text and use their hypotheses as the target text to fine-tune a BART model. Similarly, to generate the hypotheses for different classes, we append the class labels at the end of the unlabeled premises before using them as the input to the fine-tuned BART model.  

A comparison between \textsc{DBST + N} and \textsc{DBST + N$_{SB}$} on MNLI-6K can be seen in Table \ref{table:ablation}. Clearly, the performance of the \textsc{DBST + N} is superior illustrating the necessity of class specific conditional language models. Our qualitative analysis of the generated hypotheses by the single BART model reveals that in many cases, it fails to distinguish among the appended labels at the end of premises and ends up generating the same hypothesis for all classes. In contrast, class-specific BART models are able to generate more label-relevant hypotheses by focusing on one class at a time.

\paragraph{Only Confidence Masking} In our SSL framework, we use two quality assurance filters in choosing pseudo-labeled data (see Equation \ref{eqn:selection}): a) confidence masking --- whether the confidence for the predicted pseudo-label is above a pre-defined threshold, b) label consistency --- whether the predicted pseudo-label and the synthetically assigned label match. However, in general, SSL only uses confidence masking filter for quality assurance. Therefore, to evaluate the necessity of our label consistency filter, we experiment with a version of \textsc{DBST + N} where we only use the confidence masking filter on \textsc{MNLI-6K}. We denote this approach \textsc{DBST + N$_{CM}$}. We can see in Table \ref{table:ablation} that there is a drop in performance by this model compared to the \textsc{DBST + N} model. Therefore, using the additional filter based on label consistency is beneficial in ensuring data quality.

\section{Related Work}
\vspace{-3mm}

\paragraph{Semi-supervised Learning (SSL)}
SSL has gained a lot of attention in the research community for exploiting unlabeled data to further boost the performance without any additional manual annotation effort. In general, techniques based on semi-supervised learning involve predicting the labels (hard or soft) of unlabeled data and using them in some manner for training in addition to labeled data. For example, consistency regularization \cite{sajjadi2016regularization, temporalSSL, NIPS2017_68053af2, xie2019unsupervised} aims at improving model robustness by introducing an additional loss term to minimize the distance between the probability distribution predicted by the model for an unlabeled sample and its perturbed version. Self-training is a form of SSL that assigns pseudo-labels to unlabeled samples to be used for training in addition to labeled data. Self-training has seen wide applications in various NLP and machine learning tasks. To name a few, it has been used to boost the performance in image classification \cite{yalniz2019billion, xie2020self}, sentiment analysis \cite{becker-etal-2013-avaya}, conversation summarization \cite{chen2021simple}, domain adaptation \cite{Zou_2018_ECCV, pmlr-v119-kumar20c, NEURIPS2021_c1fea270} and few-shot text classification \cite{NEURIPS2020_f23d125d}. However, to date, none of these SSL methods have been used to improve the performance in low-resource NLI. 

\paragraph{Data Augmentation (DA)} DA is also a popular technique for automatically increasing the amount of labeled data used for training. However, in general, unlike semi-supervised learning, unlabeled data is not utilized in data augmentation. Rather, the labeled samples are perturbed in different manners to create variations of the same input. Simpler data augmentation approaches use rule-based methods such as synonym replacement \cite{zhang2015character}, random swap/deletion/insertion of tokens \cite{wei2019eda} etc. More complex methods use different generative models to synthesize a new version of the input text using variational autoencoders \cite{kingma2013auto}, pre-trained language models \cite{kumar2020data} and back-translation \cite{yu2018fast}. Similar to SSL, DA is yet to be explored in the context of NLI.

\paragraph{Synthetic Data Generation} With the advancement of text generation models in recent years \cite{radford2019language, brown2020language, lewis-etal-2020-bart}, researchers have started to adopt synthetic text generation as a data augmentation strategy for various tasks such as commonsense reasoning \cite{yang-etal-2020-generative}, sentence classification \cite{anaby2020not} and question answering \cite{puri-etal-2020-training}. More recently, synthetic text generation methods have also been employed to remove spurious correlations in human annotated data for fact verification \cite{Lee2021CrossAugAC} and NLI \cite{wu-etal-2022-generating}. However, to our knowledge, no prior work aims at tackling low-resource NLI scenarios nor do they address the unavailability of hypotheses for unlabeled premises by using synthetic data generation methods.

\vspace{1.5mm}
\section{Conclusion \& Future Work}

In this paper, we propose a novel method for leveraging unlabeled data for NLI in low-resource settings to improve the performance without the necessity of additional human annotation effort. We develop an SSL framework based on this proposed method for NLI which substantially improves both in-domain and out-of-domain performance of four NLI datasets in low resource scenarios illustrating that our SSL framework considerably improves the generalization capability of the models by harnessing unlabeled data. Our results indicate that we successfully address the key bottleneck in exploring SSL for NLI, i.e., the unavailability of hypothesis for unlabeled premises. Our future work will include developing methods which can improve the robustness of the models further.

\section*{Acknowledgements} 
This research is supported in part by NSF CAREER
award \#1802358, NSF CRI award \#1823292, NSF IIS award \#2107518, and UIC Discovery Partners Institute (DPI) award.
Any opinions, findings, and conclusions expressed
here are those of the authors and do not necessarily
reflect the views of NSF or DPI. We thank AWS for computational resources. We also thank our anonymous
reviewers for their constructive feedback.

\typeout{}
\bibliography{anthology, custom}
\bibliographystyle{acl_natbib}
\clearpage
\appendix

\setlength\dashlinedash{0.2pt}
\setlength\dashlinegap{1.5pt}
\setlength\arrayrulewidth{0.3pt}
\begin{table*}
\centering
  \begin{tabular} {p{16em}  p{16em} p{6em}}
    \toprule
  \multicolumn{1}{c}{\rule{0pt}{2ex}\textbf{Unlabeled Premise}} & \multicolumn{1}{c}{\textbf{Synthetic Hypothesis}}&
  \multicolumn{1}{c}{\rule{0pt}{2ex}\textbf{Synthetic Label}}\\[1ex] 
    \hline
    Started the moment I got the wire. &  It began the moment I got the wire. & \textsc{Entailment}\\
    \hdashline
    Started the moment I got the wire. & It started to happen after I pulled the wire from the car. & \textsc{Neutral}\\
    \hdashline
    Started the moment I got the wire. & It started before I got the wire. & \textsc{Contradiction}\\
    \hline
    He consolidated power at Thebes during his short reign. &  Thebes came to control power during his short reign. & \textsc{Entailment}\\
    \hdashline
    He consolidated power at Thebes during his short reign. & He was a megalomaniac obsessed with power, acquiring vast estates. & \textsc{Neutral}\\
    \hdashline
   He consolidated power at Thebes during his short reign. & He had no power at Thebes. & \textsc{Contradiction}\\
   
   \hline
    She was born in North Carolina and went to high school in Westbury, N.Y. &  Westbury, N.Y, is where she was born, and graduated from high school. & \textsc{Entailment}\\
    \hdashline
    She was born in North Carolina and went to high school in Westbury, N.Y. & Her parents did not approve of her choice of education. & \textsc{Neutral}\\
    \hdashline
   She was born in North Carolina and went to high school in Westbury, N.Y. & She is from New Jersey, where she attended college. & \textsc{Contradiction}\\

    \bottomrule
  \end{tabular}
  \caption{Examples from the synthetic dataset generated for MNLI-6K.}
    \label{table:synthetic_examples}
\end{table*}

\section{Details about Baselines}

\subsection{Conditional Masked Language Modeling (C-MLM)}
\label{appndx: C_mlm}
We use the following templates for each class to combine the premise and hypothesis of each sample in the labeled dataset along with their label information: 
\begin{itemize}
    \item \textsc{Entailment}: <\textit{Premise}> implies <\textit{Hypothesis}>.
    \item \textsc{Contradiction}: <\textit{Premise}> contradicts <\textit{Hypothesis}>.
    \item \textsc{Neutral}: <\textit{Premise}> neither implies nor contradicts <\textit{Hypothesis}>.
\end{itemize}
We mask half of the common tokens in each premise-hypothesis pair. For example, if there are 2 common tokens between a premise and its hypothesis, only one of them selected randomly and masked in both sentences.

After the tokens are predicted, we perform a post-processing step to ensure that the masked positions where the original premise and hypothesis had the same token also get replaced with the same predicted token. This was done to reduce the possibility of accidentally changing the label.

\subsection{Filter for ST  with Randomly Chosen Hypothesis}
Since, there is no synthetically assigned label for examples when the hypothesis is selected using random sampling, we could not use the label consistency filter for our \textsc{ST - RH} baseline. Specifically, equation \ref{eqn:selection} is updated as follows for this model.

\begin{multline}
\label{eqn:selection_random}
 D_{k}^{p} = \{(p_{i}^{u}, h_{i}^{u}, y_{i}^{p}) :  \\ {\theta_{k}^{clf}}( y_{i}^{p}| p_{i}^{u}, h_{i}^{u}) \ge \tau \}_{i=1,..S} 
\end{multline}

\section{Implementation Details}
\label{apndx:implementation_details}
Our implementation details can be divided into two parts: hypothesis generation and self-training.

\subsection{Hypothesis Generation}
We implement our hypothesis generation module using the huggingface transformers\footnote{\url{https://huggingface.co/docs/transformers/index}} library. Specifically, we choose `bart-large' as our hypothesis generation model for all our datasets. We fine-tune each model for 30 epochs with a cross-entropy loss. We use the AdamW optimizer \cite{loshchilov2018decoupled} with an initial learning rate of $3e-5$ and a batch size is set at 64. 
After the BART models are trained, we generate the hypotheses by using the unlabeled premises as the inputs. We employ top-$k$ sampling with $k=10$ and temperature scaling with temperature = $2.0$ to introduce diversity to the generated hypotheses.

Each hypothesis generation model was trained in $\approx 1$ hour using a single NVIDIA RTX A5000 GPU. It took $\approx 2$ hours to generate the synthetic hypotheses for each dataset using the same GPU. 

\subsection{Self-training}

At each iteration, we fine-tune a pre-trained `bert-base-cased' \cite{devlin-etal-2019-bert} model as our classifier using the huggingface transformers library where we concatenate the premise and hypothesis with a \texttt{[SEP]} token between them to be used as the input. We then project the hidden representation of the \texttt{[CLS]} token with a weight matrix $\mathbf{W} \in \mathbb{R}^{d \times |C|}$ to get the classification output. Here, $C$ is the set of classes. We fine-tune the model for $10$ epochs with a cross-entropy loss using the Adam optimizer \cite{kingma2015adam} with an initial learning rate of $2e-5$. We employ early stopping with a patience size 2 where we use the Macro F1 score of the development set as the stopping criteria. The batch size is set at 64.

We set the highest number of iterations, $K = 100$ for all our self-training experiments. However, we stop the experiments if the development score does not improve for 10 consecutive iterations. The confidence threshold, $\tau$ for \textsc{SICK}, \textsc{MNLI-6K} and \textsc{SNLI-6K} is set at $0.9$. For, \textsc{RTE}, we set $\tau = 0.7$ because higher threshold for this dataset results in very few pseudo-labeled examples getting selected. The number of synthetically labeled data to be sampled, $S$ at each iteration is set to $ \approx 0.75 * n$ where $n$ is the size of each labeled dataset. For our noisy models, we use a transformer based model\footnote{\url{https://huggingface.co/Helsinki-NLP/opus-mt-en-ROMANCE}} for back-translating both premise and hypothesis.

The self-training models took $\approx 12 - 24$ hours to finish running on a single NVIDIA RTX A5000 GPU.

\subsection{Out-of-domain Performance Evaluation for RTE}
\textsc{RTE} is a 2-way classification dataset (\textsc{entailment} and \textsc{not\_entailment}) whereas the other datasets have 3 classes (\textsc{entailment}, \textsc{contradiction} and \textsc{neutral}). Thus, for out-of-domain performance evaluation of the models trained with RTE, we convert the \textsc{contradiction} and \textsc{neutral} class labels of the other datasets to \textsc{not\_entailment}. We follow the same method for evaluating the models trained on RTE on the stress set. Similarly, when we evaluate the models trained on other datasets on RTE (as the OOD dataset), we convert the predicted \textsc{contradiction} and \textsc{neutral} label to \textsc{not\_entailment}. 

\section{Examples from Synthetically Generated Data}
\label{appndx:synthetic_examples}
Table \ref{table:synthetic_examples} shows a few examples from the synthetically labeled dataset for MNLI - 6K. We can see that the BART models are able to generate meaningful hypotheses for each unlabeled premise and provide coverage for all classes.

\end{document}